\documentclass[preprint,12pt]{elsarticle}
\usepackage{amsmath}
\usepackage{graphicx}
\usepackage{epstopdf} 
\usepackage{multirow}
\usepackage{times}
\usepackage{latexsym}
\usepackage{amssymb}
\usepackage{hyperref}
\usepackage{makecell}
\usepackage[T1]{fontenc}
\usepackage{bm}
\usepackage[utf8]{inputenc}
\usepackage{subfigure}
\usepackage{parskip}
\usepackage{microtype}
\usepackage{inconsolata}
\usepackage{booktabs}

\usepackage{amssymb}

\journal{Journal of Biomedical Informatics}

\begin{document}

\begin{frontmatter}



\title{\textsc{W-procer}: Weighted Prototypical  Contrastive Learning for  Few-Shot Medical Named Entity Recognition}


\author[inst1]{Mingchen Li}

\affiliation[inst1]{organization={Division of Computational Health Sciences, Department of Surgery, University of Minnesota},
            city={Minneapolis},
            state={MN},
            country={USA}}

\author[inst2]{Yang Ye}

\affiliation[inst2]{organization={Department of Computer Science, Georgia State University},
            city={Atlanta},
            state={GA},
            country={USA}}
            
\author[inst1]{ Huixue Zhou}
\author[inst1]{Jeremy Yeung}
\author[inst2]{Huaiyuan Chu}
\author[inst1]{Rui Zhang\footnote{Corresponding author at: 8-100 Phillips-Wangensteen Building, 516 Delaware Street SE, Minneapolis, MN 55455, University of Minnesota.
E-mail address: zhan1386@umn.edu (R. Zhang).}}

\begin{abstract}
Objective: Contrastive learning (CL) has become a popular solution for few-shot Name Entity Recognition (NER). Nevertheless, existing CL methods result in a widening gap between labeled entities and \textsc{Outside (O)} labeled tokens, while at the same time, they bring entities with the same label closer together,
disregarding the fact that the relevant entities of these labeled entities are already labeled as \textsc{Outside (O)}.  Therefore, it is not prudent to merely extend the distance between labeled entities and \textsc{Outside (O)} labeled tokens.
To address this challenge, we propose a  novel method named \textbf{W}eighted \textbf{Pro}totypical  \textbf{C}ontrastive Learning for  Few-Shot Medical Named \textbf{E}ntity \textbf{R}ecognition (\textsc{W-Procer}).
 
Methods:  Our approach primarily revolves around constructing the prototype-based contrastive loss and weighting network. These components play a crucial role in assisting the model in differentiating the negative samples from \textsc{Outside (O)} labeled tokens and enhancing the discrimination ability of the model.
 
Results: Experimental results show that our proposed \textsc{W-Procer} framework significantly outperforms the strong baselines on the three medical benchmark datasets.
For instance, notable enhancements of approximately 3.3\% in Micro-F1 for the NCBI 5-shot dataset and around 3.5\% for the I2B2'14 5-shot dataset have been accomplished in the NER task.

Conclusion: The obtained results lead us to the conclusion that our proposed weighted prototypical contrastive learning techniques are well-suited for the medical few-shot Named Entity Recognition (NER) task. These findings underscore the significance of employing type-based contrastive loss and prototype-based contrastive loss to enhance the identification of medical entities. The explored techniques not only showcase their robustness in addressing the class collision problem but also exhibit their applicability to other medical NLP tasks facing similar challenges.
\end{abstract}

\begin{graphicalabstract}
\includegraphics[width=1\columnwidth]{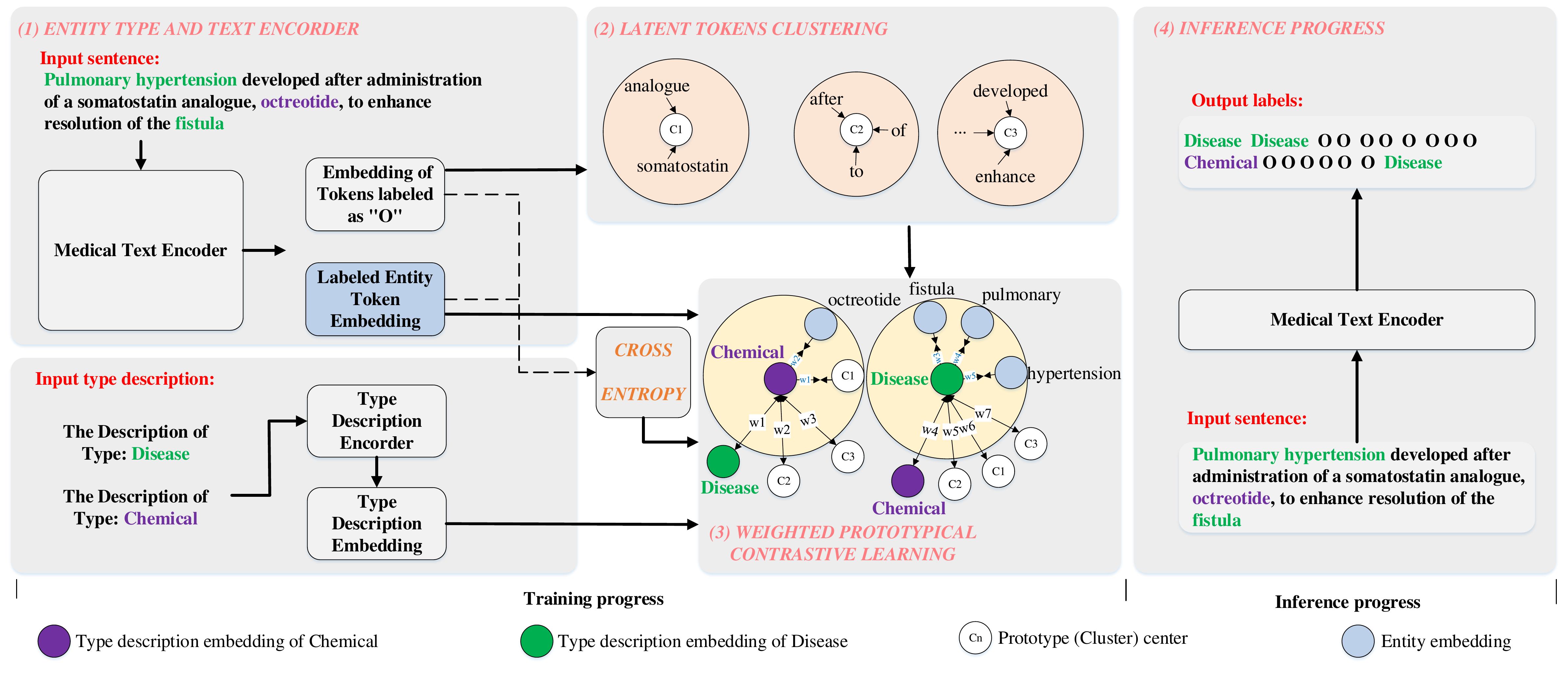}
\end{graphicalabstract}


\begin{highlights}
\item We propose  \textsc{W-Procer}, a new framework that uses weighted prototypical contrastive learning to solve the class collision problem in the few-shot medical NER  task.

\item  We evaluate the effectiveness of our weighting strategies in the negative samples.

\item We conduct a thorough analysis of our method, including an ablation study, demonstrating the effectiveness of  \textsc{W-Procer}.
\end{highlights}

\begin{keyword}
Contrastive Learning \sep Few-Shot Medical
Named Entity Recognition \sep Language Model
\end{keyword}

\end{frontmatter}



\section{Introduction}
Researchers are increasingly interested in applying information extraction to mine a vast quantity of unstructured information from electronic medical records.
These techniques can offer valuable perception  and generate substantial benefits  for 
clinical research, such as drug discovery \cite{zhang2021drug}, and knowledge graphs building~\cite{li2020multi,wu2023medical}. Within the scope of medical text mining, one of the  most essential tasks in medical text mining is medical named entity recognition (NER). However,
 existing supervised medical NER models necessitate a substantial amount of human-annotated (i.e. medical student, doctor) data. To tackle this issue, few-shot techniques have been introduced to perform NER in resource-constrained settings by leveraging auxiliary information or improving the discrimination between different labels\footnote{In this paper, the terms "type" and "label" are used interchangeably to convey the same meaning.}.

Few-shot learning~\cite{yang2020simple,das2021container,li2022hierarchical}  consists of learning seen or unseen classes from insufficiently labeled data. To address the challenges posed by the scarcity of available medical data,
several studies~\cite{hou2020few,guibon2021few,das2021container,huang2022copner}  propose to harness the power of contrastive learning~\cite{chen2020simple}.
 Nonetheless, these studies share a common limitation in the NER task. Unlike the entity type that carries a clear semantic representation, the tokens with \textsc{Outside (O)} label lack a unified semantic meaning. 
In simpler terms, the training data contains only partial annotations~\cite{koch2017multi}, significantly reducing the effort required for labeling each medical sentence. Consequently, this allows expert raters to partially annotate a larger number of training NER sentences in the same timeframe.
This issue becomes increasingly pronounced in fields like medicine where there is an abundance of labels.
So, in the current CL-based methods, the operation of reducing the distance between tokens with the same label (positive pair) and increasing the distance between tokens with different labels (negative pair) inevitably yields negative pairs that share similar semantic label meanings and should be closer in the embedding space. 
Therefore the model will struggle to predict the correct labels in the test set.  
This problem is defined as a class collision in \cite{arora2019theoretical} and is shown to hurt  representation learning in \cite{li2020prototypical}. 
For instance, in the sentence \textit{...after administration of a somatostatin analogue, octreotide,  to enhance,..} from  dataset BC5CDR~\cite{li2016biocreative}, \textit{octreotide} is labeled as \textit{"Chemical"}, other tokens are labeled as \textit{"O"}.
Considering \textit{somatostatin, analogue} and \textit{octreotide} as the negative pairs  would be unreasonable because of their relevance to the type \textit{Chemical}.

To address the above issue, we aim to develop an enhanced CL method that mitigates the impact of the class collision in the few-shot medical NER task.
The core idea is motivated by PCL~\cite{li2020prototypical} from unsupervised image representations, which  involves constructing multiple prototypes from sampled examples and designing a contrastive loss function that ensures the embedding of a given sample is more similar to its corresponding prototypes as opposed to other prototypes. Thus we propose a novel framework towards \textbf{W}eighted  \textbf{Pro}totypical \textbf{C}ontrastive Learning for  Few-Shot Medical Named \textbf{E}ntity \textbf{R}ecognition, namely \textsc{W-Procer}.

In our approach, we first construct several prototypes from  \textsc{Outside (O)} labeled tokens by the cluster method, such as K-means~\cite{li2012clustering}, and define the distance between the observed data and the center of clustered prototypes to help distinguish the positive examples and negative examples from the \textsc{Outside (O)} data.
Next, we construct a prototype-based contrastive learning that ensures the embedding of a given sample is more similar to its corresponding prototypes as opposed to other prototypes.
Additionally, we introduce a type-based contrastive loss that encourages the reduction of distances between entities with the same type. To facilitate the model's ability to distinguish the importance of various negative samples, we propose a weighting network that is utilized in both of the aforementioned contrastive objectives.
We perform extensive experiments on three standard medical datasets for few-shot
NER demonstrates the effectiveness of our method over prior state-of-the-art methods. 



\section{Related Work}

\subsection{Few-Shot NER}
Few-shot named entity recognition is a task that aims to predict the label (type) of an entity from insufficient labeled data. Most previous work in the medical domain prefer to directly use medical language models such as ClinicalBERT~\cite{huang2019clinicalbert}, BioBERT~\cite{lee2020biobert}, and GatorTron~\cite{yang2022large}  to solve the few-shot NER problem.  A few studies
\cite{fritzler2019few, ji2022few} propose to utilize the prototype network \cite{snell2017prototypical} to catch the few-shot medical NER tasks. 
Inspired by the  nearest neighbor inference~\cite{wiseman2019label}, \cite{yang2020simple} proposes NNshot and Structshot, which use the nearest neighbor to search for the  nearest label of each testing entity.
In CONTaiNER~\cite{das2021container}, the authors use contrastive learning to increase the  discrimination for each label and adopt Gaussian embeddings for each token to solve the Anisotropic  property in the few-shot NER task. 
The prompt-based method is also explored in this task, such as \cite{huang2022copner} uses the prompt to guide contrastive learning. 
\cite{zhang2022optimizing} proposes a span-based contrastive learning method to handle the nested NER.
Although performance has been achieved incrementally, these approaches  largely ignore the class collision, which undermines the predictive performance during label prediction. To mitigate this issue, \textsc{W-Procer}
first constructs the  prototype from the \textsc{Outside (O)} labeled tokens, which is then used to reduce the class collision by proposed weighted prototypical contrastive learning.

\subsection{Contrastive Learning}
Many studies~\cite{li2020prototypical,chen2020simple,dangovski2021equivariant,suresh2021not,das2021container,zhang2022optimizing,huang2022copner,chuang2022diffcse} have been proposed to enhance the discrimination for different labels and improve the robustness of the sample representation. For example, on the image representation  task, 
SimCLR~\cite{chen2020simple} first proposes a contrastive learning  method to  enhance image representations. Different from the learning strategy of SimCLR~\cite{chen2020simple}, \cite{li2020prototypical} regards there are many negative
pairs  that share similar semantics, it is inappropriate to push them from each other in the embedding space, so they propose prototypical contrastive learning to ensure the embedding of a given sample is more similar to its corresponding prototypes as opposed to other prototypes. On the  text representation task,  \cite{suresh2021not} proposes a weighting network to help differently weight the positives and negatives in the  contrastive objective function, while DiffCSE~\cite{chuang2022diffcse} introduces a strategy that employs  equivariant contrastive learning to learn the insensitive and sensitive features separately.
Inspired by these studies, we have developed an innovative prototype-based CL and weighting network to enhance the accuracy of entity type prediction in the field of medical few-shot NER.


\section{Methodology}
\begin{figure*}[htbp]
	\centering
	\includegraphics[width=1\columnwidth]{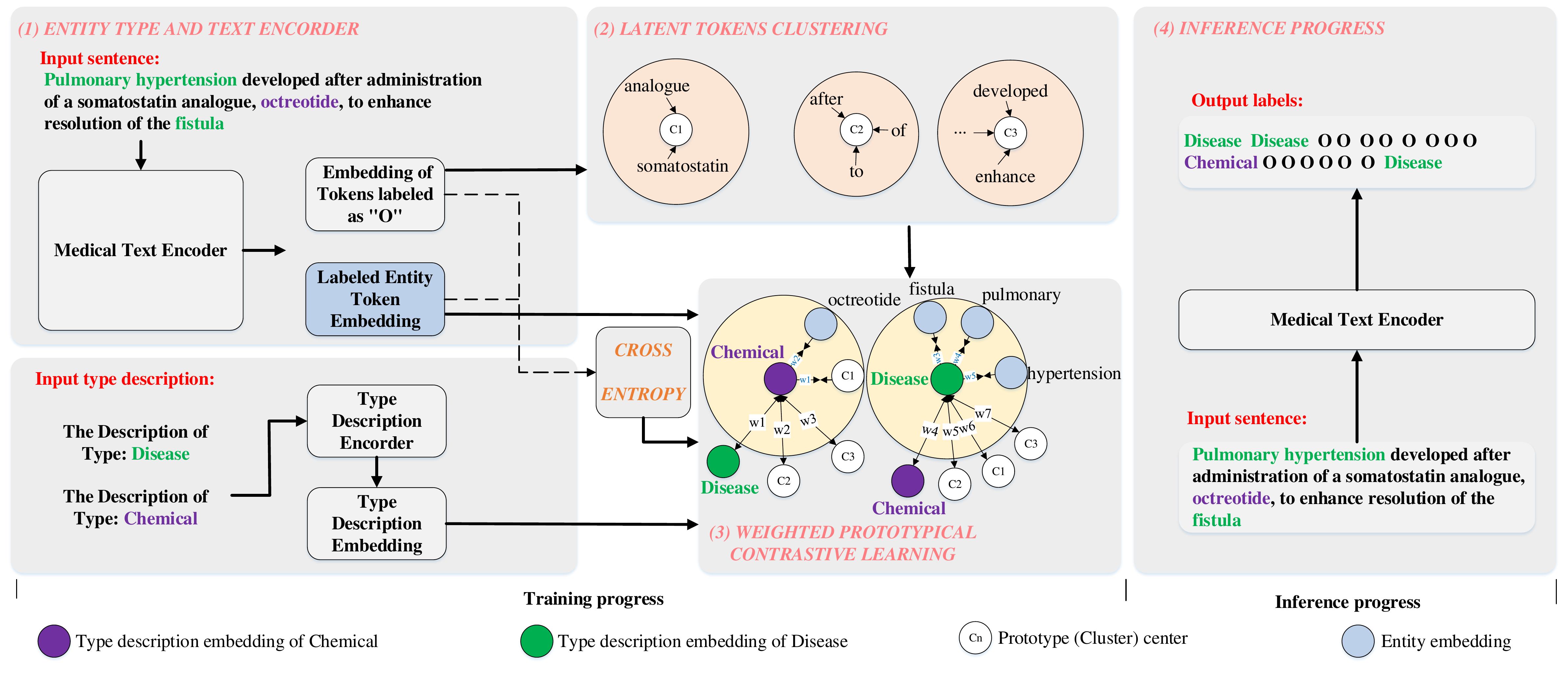} 
	\caption{Training progress and Inference progress of \textsc{W-Procer}}
	\label{con:PWBFB}
\end{figure*}
In this paper, we propose  a novel model  which uses the weighted prototypical contrastive learning to improve the performance of few-shot medical NER.
Figure~\ref{con:PWBFB} gives an overview of our \textsc{W-Procer}, which consists of four main steps: (1) it encodes the input sentence and the description of class type into the sentence embedding and type embedding by the \textbf{Entity Type and Text Encoder Moudle}, here we use GatorTron~\cite{yang2022gatortron} as the medical knowledge encoder; (2) it  clusters the token embeddings of labeled \textsc{Outside (O)} in the (support) training dataset and  constructs the positive prototype center and negative prototype center  in the \textbf{Latent tokens Clustering} module; (3) it incorporates the tokens embedding, entity  type embedding and the centers of the different prototypes  into \textbf{Weighted Prototypical Contrastive Learning} (WPCL) to help model differentiate the   entities with different labels and the negative samples from \textsc{Outside (O)} labeled tokens. After that, the model is optimized by the  WPCL and the Cross-Entropy; (4) 
we predict the label sequence with the maximum score, this progress is shown in \textbf{Inference Progress}. Next, we provide details for each component.

\subsection{ Entity Type and Text  Encoder}
Entity type embeddings are regarded as the anchors in the vector space.  For the different types, we first define their textual definition based on Wikidata\footnote{https://www.wikidata.org}
and calculate their embeddings using the medical language model (MLM)  GatorTron with an additional linear projection. For later use, we define the embeddings of the set  of entity type descriptions, $T=\{t_1,...,t_n\}$ as  $\mathbf{T}=\{\mathbf{t_1},...,\mathbf{t_n}\}$,  $\mathbf{t_n}$ is the embedding for the entity type description  $t_n$.  Similar to the embedding of entity type  descriptions, we again use GatorTron as the text encoder and utilize  the final hidden states to represent the basic text tokens in the support set,

\begin{align}
& \{\mathbf{h_1},\mathbf{h_2},...,\mathbf{h_l}\} =MLM \{x_1,x_2,...,x_l\} \label{sentence_represenation}
\end{align}
where $l$ is the length of tokens in an input sentence.

\subsection{Latent Tokens Clustering}
We perform a latent clustering strategy on \textsc{Outside (O)} labeled tokens from the training dataset. This is based on the fact that there are many unannotated tokens in the  \textsc{Outside (O)} labeled tokens.
Then we set the threshold $\alpha$ to  judge the positive pair between different prototype centers and anchors.
Specifically, we perform token clustering on these \textsc{Outside (O)} labeled tokens  $\{\mathbf{o_1},\mathbf{o_2},...,\mathbf{o_j}\}$ to obtain $k$ clusters in a sentence. We define the prototype center $c_k$ as the centroid embedding for the $k$-th cluster. Then we construct the  positive set $P^c_n$ and the negative set $N^c_n$ for the anchor $\mathbf{t_n}$  by,

\begin{align}
\begin{cases}
c_k \in P^c_n & D(c_k,\mathbf{t_n})<\alpha \\
c_k \in N^c_n & D(c_k,\mathbf{t_n})>\alpha
\end{cases}
\label{con: distance_D_calcuation}
\end{align}
By calculating the distance between $c_k$ and $\mathbf{t_n}$, if $t_n$ belongs to the $k$-th cluster, $c_k$ will be regarded as an element in the positive set  $P_n^c$. In our work, the distance is calculated by $D$ (Euclidean distance).

\subsection{Weighted Prototypical Contrastive Learning}
\label{Prototypical Weighted Contrastive Learning}
The traditional contrastive loss in the few-shot NER task aims to reduce the distance between the semantic representations of tokens that belong to the same type by defining a set of positive pairs while increasing the distance between the semantic representations of tokens belonging to different  types by defining a set of negative pairs. 
The positive set $P_i$ for a given entity token $x_i$ about label $y_i$ only has one sample, the argument version  of $x_i$ is defined as $g(x_i)$,
which could either be the entity token sharing the same label class with token $x_i$ or the type description relating to the label type of $x_i$.
The negative set $N_i$ would be the tokens that have different label classes with $x_i$ or the type description of other labels and the  \textsc{Outside (O)} labeled tokens.
The contrastive loss can be defined as:

\begin{small}
\begin{align}
& L_{cl}=\sum_{i=1}^{K} -log \frac{exp(x_i \cdot g(x_i)/ \tau)}{\sum_{n(x_i) \in N_i}^{} exp(x_i \cdot n(x_i)/ \tau)}  \label{traditionl_CL_0}
\end{align}
\end{small}

where $K$ is the number of entity tokens, $K$ is  smaller than $l$. $\tau$ is the temperature hyper-parameter. Larger  $\tau$ will decrease the dot-products, creating a more different comparison.

From Eq. (\ref{traditionl_CL_0}), we  find  it faces one fundamental weakness, all tokens labeled as \textsc{Outside (O)} can be easily considered as elements in the negative set.
To address this, we first introduce a type-based contrastive loss to differentiate tokens with different types. Subsequently, we propose a prototype-based contrastive loss to identify positive and negative prototypes within the \textsc{Outside (O)} labeled tokens for the anchor $\mathbf{t_n}$. They are defined as:

\begin{small}
\begin{align}
& L_{type}=\sum_{i=1}^{K} -log \frac{  exp(\mathbf{t_n} \cdot \mathbf{h_i^n}/ \tau)}{\sum_{\mathbf{t_m} \in \mathbf{T}, m\neq n}^{} exp(\mathbf{t_n} \cdot \mathbf{t_m})/ \tau)} \label{function1_no_weight}
\end{align}
\end{small}
\begin{small}
\begin{align}
& L_{prototype}=\sum_{s=1}^{M} -log (\frac{\sum_{c_k \in P^c_n}^{} exp(\mathbf{t_n} \cdot c_k/ \tau)}{\sum_{c_v \in N^c_n}^{} exp(\mathbf{t_n} \cdot c_v)/ \tau)})_s \label{function2_no_weight}
\end{align}
\end{small}

where $\mathbf{t_n}$ is the embedding of the type class $t_n$, $\mathbf{h_i^n}$ is the embedding of the entity token $x_i$ with the type class $t_n$. $M$ is the number of sentences in a batch.

Eq. (\ref{traditionl_CL_0}) weights all negative samples equally to the current entity token $x_i$.  However, not all negatives are equal.  For example,  we assume the type $t_n$ is \textit{DiseaseClass},  \textit{DiseaseClass} and type \textit{SpecificDisease} are conceptually closer than \textit{DiseaseClass} and class type \textit{CompositeMention}. The same situation occurs in  our type or prototype-based  contrastive loss.
Thus our goal was to introduce a method  that can  adaptively weight the negative pair, thereby helping the model differentiate the more difficult negatives and improve the discrimination ability of contrastive learning. So the $L_{type}$ and $L_{prototype}$  are redefined as:

\begin{small}
\begin{align}
& L_{type}=\sum_{i=1}^{K} -log \frac{  exp(\mathbf{t_n} \cdot \mathbf{h_i^n}/ \tau)}{\sum_{\mathbf{t_m} \in \mathbf{T}, m\neq n}^{} w_{mn} \cdot exp(\mathbf{t_n} \cdot \mathbf{t_m})/ \tau)} \label{function1_no_weight}
\end{align}
\end{small}
\begin{small}
\begin{align}
& L_{prototype}=\sum_{s=1}^{M} -log (\frac{\sum_{c_k \in P^c_n}^{} exp(\mathbf{t_n} \cdot c_k/ \tau)}{\sum_{c_v \in N^c_n}^{} w_{nv} \cdot exp(\mathbf{t_n} \cdot c_v)/ \tau)} )_s\label{function2_no_weight}
\end{align}
\end{small}

Here, $ w_{mn}$ indicates the relationship between the entity type $t_n$ and entity type $t_m$, $w_{nv}$ indicates the relationship between the entity type $t_n$ and the negative prototypical center $c_v$. 

In our contrastive loss function, we want to increase the weight of confusable negative examples relative to other negative labels. To weight each comparison sample differently, we use two different weighting networks for  Eq.(\ref{function1_no_weight}) and Eq. (\ref{function2_no_weight}).  The prediction probabilities obtained from the softmax layer are given by:

\begin{small}
\begin{align}
& W_1= softmax (\frac{\mathbf{T}\mathbf{T}^\mathrm{T}}{\sqrt{d_k}}), w_{mn} \in W_1  \nonumber \\
& W_2= softmax (\frac{\mathbf{[T:C]}\mathbf{[T:C]}^\mathrm{T}}{\sqrt{d_k}}), w_{nv} \in W_2
\end{align}
\end{small}
$:$ denotes the concatenate operation. $\mathbf{C}$ refers to the embedding matrix of all cluster centers. 

In contrast to previous contrastive learning methods in NER, such as \cite{das2021container} that push tokens labeled as \textsc{Outside (O)}  towards all other type-labeled entities, it enables a better understanding of which tokens are labeled as \textsc{Outside (O)} in the training dataset. In our work we also need a strategy to let the model have the ability to distinguish which prototypes should be labeled as \textit{O}.
So we adopt a direct approach where we minimize the distance between the input sentence and its gold labels using cross-entropy during the training process. This allows the model to learn the mapping between entity types and their corresponding tokens labeled in the training set, and then we use our proposed CL loss functions which helps the unlabeled entities within the \textsc{Outside (O)}  tokens to converge towards their most relevant entity types. Intuitively, these two steps have conflicts. 
On one hand, we aim to minimize the distance between the positive prototype derived from \textit{O} labeled tokens and \textit{O} label itself. On the other hand, we aim to minimize the distance between a positive prototype and its relevant type anchor.
By implementing this operation, we maintain the assumption that the model is capable of distinguishing token labels between the \textit{O} label and entity types. The experiments demonstrate that our model has a higher tendency to align the positive prototype with its relevant type anchor.
Our overall objective is as follows:

\begin{small}
\begin{align}
& L_=L_{CE}+\beta L_{type} +(1-\beta)L_{prototype}  \label{final_loss_function}
\end{align}
\end{small}

Where $L_{CE}$ is the cross-entropy loss, which optimizes the output of the last hidden layer and the entity types labeled in the training data.   $\beta \in [0,1]$ is a tunable parameter.

\subsection{Inference Progress}
During the inference, given the input token $x_i$ and the candidate entity types  $\{L^n\}$, the label $y_i$ for token $x_i$ is predicted at the final layer of MLM with the maximum score:

\begin{small}
\begin{align}
&\{y1,y2,...,y_i\}=\mathop{argmax}_{y^{'}  \in \{L^n\}} p (y^{'} | \{\mathbf{h_1},\mathbf{h_2},...,\mathbf{h_i}\})
\end{align}
\end{small}

\section{Experiments}

\subsection{Dataset}
\textsc{W-Procer}  is evaluated with three medical datasets, including I2B2'14~\cite{stubbs2015annotating}, BC5CDR~\cite{li2016biocreative} and NCBI~\cite{dougan2014ncbi}.  Among these datasets,  NCBI and BC5CDR consist of 798 and 1500 public medical abstracts separately, all of which are annotated with MeSH identifiers.  The data I2B2'14 originates from  the 2014 i2b2 Challenge and is provided by Partners HealthCare. All records have undergone complete de-identification and have been manually annotated to identify risk factors associated with diabetes and heart disease. The dataset I2B2'14 has 23 entity types, 140,817 sentences, and 29,233 entities.  The dataset BC5CDR has 2 entity types, 13,938 sentences, and 28,545 entities. The dataset NCBI has 4 entity types, 7,287 sentences, and 7,025 entities.

\subsection{Baselines and Evaluation Metrics}
We compare the performance of \textsc{W-Procer} with several strong baselines based on the state-of-the-art few-shot NER models, including both Domain Transfer and Domain non-Transfer methods. Specifically, the Domain Transfer models have been trained on the OntoNotes 5.0 dataset, which comprises medical NER knowledge. The evaluation of these models is conducted using the support sets and test sets from  I2B2'14, BC5CDR, and NCBI.  On the contrary, Domain non-Transfer models differ in that they do not undergo additional pre-training using other medical NER datasets.

We consider the following Domain Transfer models:  (1) \textbf{NNshot}\cite{yang2020simple} is a method that uses nearest neighbor classification;
(2) \textbf{Structshot}\cite{yang2020simple} is an improved version of NNshot that combines nearest neighbor classification, abstract transition matrix, and Viterbi algorithm; (3) \textbf{ContaiNER}~\cite{das2022container}  adopts contrastive learning to estimate the distributional distance between entities' vectors, which are represented using Gaussian embeddings; (4) \textbf{COPNER}~\cite{huang2022copner} leverages contrastive learning with prompt tuning to  identify entities;  (5) \textbf{EP-NET}~\cite{ji2022few} is a NER method based on the dispersedly distributed prototypes network.

We also consider the following Domain non-Transfer models: (1) \textbf{ProtoBERT}~\cite{huang2022copner} is a few-shot NER method, which utilizes the prototypical network~\cite{snell2017prototypical} and BERT model to infer the entity label. (2) \textbf{BINDER}~\cite{zhang2022optimizing} employs span-based contrastive learning to effectively push the entities of different types by optimizing both the entity type encoder and sentence encoder.
(3) \textbf{LM-tagger} (BERT~\cite{devlin2018bert}, ClinicalBERT~\cite{huang2019clinicalbert}, BioBERT~\cite{lee2020biobert} and GatorTron~\cite{yang2022large}) are traditional LM-based methods  that fine-tune the LM on the support set with the label classifier. 

Same as ~\cite{yang2020simple,das2022container,huang2022copner,ji2022few,zhang2022optimizing}, we evaluate all the models based on the generative evaluation metric, Micro F1.

\subsection{Settings}
In \textsc{W-Procer}, during training, we use AdamW~\cite{loshchilov2017decoupled} as the optimizer and our Contrastive loss with Cross-Entropy as the loss function with a learning rate of $0.00005$. We use label smoothing~\cite{muller2019does} to prevent the model from being over-confident. 
For the overall loss function in Eq. (\ref{final_loss_function}), we tune the $\beta$ value in 0, 0.3, 0.5, 0.9, 1. We find that \textsc{W-Procer} works well when the weight of $L_{prototype}$ is 0.5.
For the number of prototype centers, we tune the threshold $k$ in $\{3,4,5\}$, and when $k=3$, our approach achieves the best performance.   For the distance to judge the positive 
prototype center and negative center, we tune the threshold $\alpha$ in $\{0.7,0.8,0.9\}$, and when $\alpha=0.7$, our approach achieves the best performance.
In this work, we utilize greedy sampling~\cite{yang2020simple} to sample  the support set, and use the original test set for our prediction.  To clarify the tagging scheme setup, we utilize the \textsc{"IO"} tagging scheme. Under this scheme, the \textsc{"I"}  indicates that all tokens are inside an entity, while \textsc{"O"} refers to all other tokens.

\subsection{Results}
\label{Results and Discussion}

\begin{table}[ht]
	\centering
 \renewcommand\arraystretch{1.3}
\resizebox{0.9\textwidth}{!}{%
	\begin{tabular} {l|l | ccc |ccc}
		\toprule 
		\multicolumn{1}{c}  {}&\multicolumn{1}{c}  {}& \multicolumn{3}{c}  {1-shot}& \multicolumn{3}{c}  {5-shot}\\
		&Approach &   I2B2'14&BC5CDR &NCBI & I2B2'14&BC5CDR&NCBI   \\ 
          \midrule
        
         \multirow{5}*{Domain Tranfer}&NNShot~\cite{yang2020simple}   &  16.60$\pm$2.10 &   32.96$\pm$ 1.67 & 11.82$\pm$ 1.02  & 23.70$\pm$1.30&39.30$\pm$1.24 & 16.22$\pm$0.69 \\
         ~&StructShot~\cite{yang2020simple}   & 22.10 $\pm$3.00 &16.09$\pm$ 1.39& 4.63$\pm$ 0.70  &  31.80$\pm$1.80&30.97$\pm$1.77& 13.89$\pm$1.09 \\
          ~&ContaiNER~\cite{das2022container}   & 21.50$\pm$1.70 & 37.25$\pm$5.52 &16.51$\pm$3.90  &   36.70$\pm$2.10& 41.21$\pm$7.31   & 26.83$\pm$3.18\\
          ~&COPNER~\cite{huang2022copner}  & 35.80$\pm$1.30 &36.36$\pm$1.85 &  15.54$\pm$1.43  & 43.70$\pm$1.50&42.78$\pm$0.34 & 24.23$\pm$1.51\\
          ~&EP-NET~\cite{ji2022few}& 27.50$\pm$4.60 &--&  -- & 44.90$\pm$2.70& -- &--\\
        \midrule
        \multirow{6}*{Domain non-Tranfer}&BINDER~\cite{zhang2022optimizing} &  9.68$\pm$5.02  &1.86$\pm$0.31& 2.95$\pm$0.71 & 32.76$\pm$6.44 & 51.79$\pm$14.56 & 31.95$\pm$1.25 \\
        ~&ProtoBERT~\cite{huang2022copner}  &  13.40$\pm$3.00 &23.61$\pm$6.80 &  17.24$\pm$1.64&  17.90$\pm$1.80 &40.58$\pm$6.17&34.18$\pm$1.02 \\
        ~&BERT~\cite{devlin2018bert} &  18.79$\pm$7.14   &10.58$\pm$4.90   &  12.33$\pm$3.14  &30.38$\pm$2.68 & 35.60$\pm$1.80  & 26.53$\pm$3.62\\
        ~&ClinicalBERT~\cite{huang2019clinicalbert} &17.46$\pm$4.11&  19.96$\pm$2.40  & 8.27$\pm$1.89 & 18.97$\pm$2.68 & 39.25$\pm$4.20 & 24.47$\pm$2.14\\ 
        ~&BioBERT~\cite{lee2020biobert}  &   11.25$\pm$0.00  & 36.78$\pm$4.29    &  27.19$\pm$1.60  & 20.50$\pm$2.07& 47.04$\pm$1.39&  35.88$\pm$2.55 \\
        
        ~&GatorTron~\cite{yang2022large} & 35.25$\pm$6.53    & 26.97$\pm$4.79   & 35.00$\pm$3.57 & 42.94$\pm$4.16 &55.44$\pm$2.43  &37.64$\pm$1.76 \\
         \midrule 
        ~&\textsc{W-Procer} w/o weight &   36.50$\pm$5.12  &38.10$\pm$5.87  & 37.19$\pm$1.47  &  45.06$\pm$3.52     & 41.59$\pm$1.68 & 39.22$\pm$2.80 \\
        ~&\textsc{W-Procer} w/o prototype & 35.41$\pm$3.91    & 37.13$\pm$4.13  & 37.93$\pm$2.14 &     45.07$\pm$1.96 & {43.83$\pm$2.34} & 40.52$\pm$3.94 \\
         ~&\textsc{W-Procer} (Our Approach) &36.91$\pm$4.46   & 40.26 $\pm$4.46  &38.86$\pm$2.80 &     46.43$\pm$4.30&  56.02$\pm$3.06 & 40.90$\pm$1.63 \\
        \bottomrule            
	\end{tabular}
 }
  \caption{Results of various approaches for 1/5-shot  medical NER on I2B2'14, BC5CDR and NCBI. 
  The baseline results of ProtoBERT, NNshot, Structshot, ContaiNER, COPNER, and EP-NET on the I2B2'14 dataset were obtained from their respective source papers. 
  For the remaining baseline models, we conducted training using their source code on the I2B2'14, BC5CDR, and NCBI datasets. It is important to note that for EP-NET, as the source code is not available, we solely report their results from the paper specifically on the I2B2'14. We incorporate PubMedBERT into BINEDR, leveraging its superior performance as demonstrated in the source paper.
  }
  \vspace{+2mm}
\label{con:Model_performance}
\end{table}

Table~\ref{con:Model_performance} presents the experimental results of various approaches based on the Micro-F1. We have the following observations: (1) Our \textsc{W-Procer} consistently outperforms all strong baseline models across all datasets. We will conduct a more in-depth analysis of this phenomenon at a later stage.
(2) Using medical LM with domain non-transfer (such as GatorTron) shows obvious superiority compared with domain transfer baselines, especially on the 5-shot of  I2B2'14, BC5CDR, and NCBI. We speculate that this is because the baselines trained with the domain transfer strategy can not contain rich medical knowledge; (3) By employing different seed values during the training of the baseline, certain models exhibit a significant standard deviation. For instance, when considering the 5-shot results of BINDER on BC5CDR, we deduce that the difficulty lies in these models' ability to effectively transfer knowledge from a limited dataset and maintain overall model stability. (4) Compared with NNShot, StructShot is not easy to identify complex medical entities by using the abstract transition, nearest neighbor, and Viterbi, we guess the reason is some abbreviated and complex medical nouns influence the judgment ability of models. For instance in NCBI, the entities \textit{PWS, UPD, RCC}.

\subsection{Discussion}
In order to further explore the effectiveness of our
framework, we perform a series of analyses based
on different characteristics of our model.  So we propose two
model variants to help us validate the advantages
of our proposed prototype-based CL and weighting network in medical NER task.  Next, we explore the performance of  our
model with a different  number of prototype centers, the weight value in Eq.~(\ref{final_loss_function}), and the  threshold to select the positive prototype.
In the last, we did the  experiments about the \textsc{W-Procer} performance in choosing different entity descriptions, the impact of different
mask strategies which are used to mask some labeled entities and the visualization of learning representation.

\subsubsection{Ablation Experiments}
The contribution of our model components can also be learned from ablated models. We introduce two ablated models of \textsc{W-Procer}, (1) \textsc{W-Procer} w/o weight  excludes the  weighting network in \textsc{W-Procer}. (2)   \textsc{W-Procer} w/o prototype uses the weighting network and type-based contrastive learning  to improve the distinction of different labels, instead of our proposed  \textsc{W-Procer} that clusters the \textsc{Outside (O)} labeled tokens, and pushes the negative  prototype center from the anchors.
We find that by clustering the \textsc{Outside (O)} labeled tokens and pushing the negative prototype center, our \textsc{W-Procer}  significantly improves over the baseline \textsc{W-Procer} w/o prototype. This is especially true for the micro-F1 value of 1/5 shot on BC5CDR, demonstrating the effectiveness of the prototype-based contrastive learning method. We also find the weighting network can help  differentiate the more difficult negatives and improve the discrimination ability of contrastive learning,  thus the micro-F1 of the approach is significantly improved, which can be seen by comparing \textsc{W-Procer}  with \textsc{W-Procer} w/o weight.

\subsubsection{The Impact of Hyper-Parameters}
\begin{table}[ht]
	\centering
  \resizebox{0.7\textwidth}{!}{%
	\renewcommand\arraystretch{1.3}
	\scalebox{0.67}{
	\begin{tabular} {c|c|c|c|c|c|c|c|c|c|c|c|c|c|c|c}
    \toprule 
		I2B2'14 No.& 1& 2&3&4&5& 6& 7&8&9&10& 11& 12&13&14&15\\ 
	   \midrule
             $k$&3 &3&3 &3&3&4 &4&4 &4&4& 5& 5&5&5&5\\
             $1-\beta$&0 &0.1&0.5 &0.7&1&0 &0.1&0.5 &0.7&1&0 &0.1&0.5 &0.7&1\\
             $\beta$&1 &0.9&0.5&0.3&0&1 &0.9&0.5&0.3&0&1 &0.9&0.5&0.3&0\\
	  \bottomrule     
	\end{tabular}}
 }
	\caption{Different $k$,   and $\beta$ settings }
\label{con: para_setting}
\end{table}

\begin{figure}[t]
        \centering
        \includegraphics[width=0.7\columnwidth]{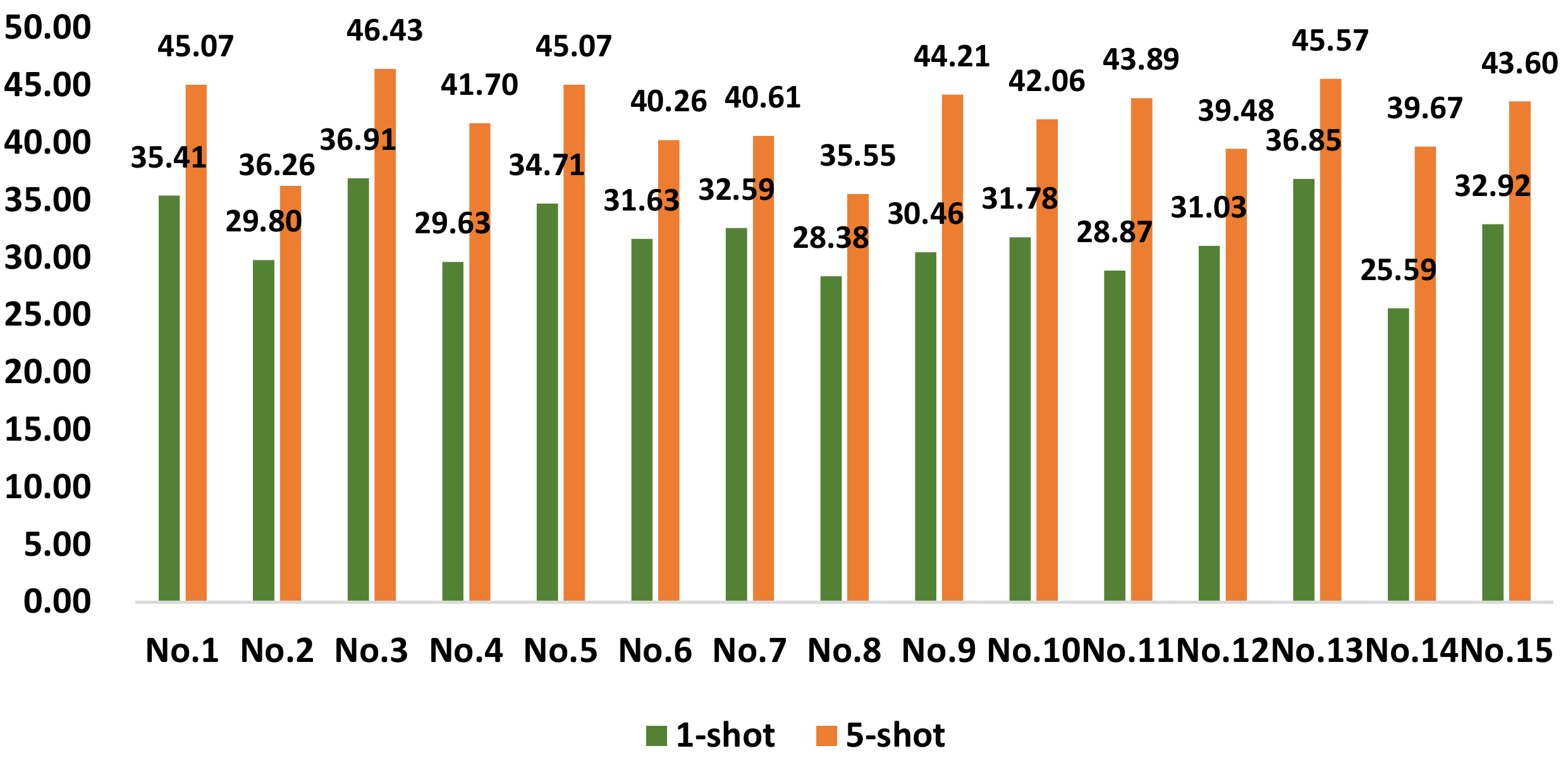}
	\caption{ Micro F1 results with different $k$, and $\beta$ settings. We fixed the distance threshold $\alpha=0.7$. }
	\label{con:parameter_setting_0}
\end{figure}

In our proposed model, there are two parameters controlling the number of prototype centers and the weight value of prototype-based CL and type-base CL: (1) $k \in \{3, 4, 5\}$ which is the cluster number to  help cluster the \textsc{Outside (O)}  based on their semantic distance, and (2) $\beta\in\{1, 0.9,0.5,0.3,0\}$ which is a tunable parameter. To investigate the influence of different $k$ and $\beta$, we conduct experiments using
\textsc{W-Procer} with different parameter settings which are shown in Table \ref{con: para_setting}.  
We analyze the impact of these two hyper-parameters in Figure~\ref{con:parameter_setting_0}. As we can see, when $k=3$, \textsc{W-Procer} tends to select prototypes  that are more relevant to the type description and consistently provides better performance than the setting of $k=4,5$ with the $\beta=0.5$.
\begin{figure}[t]
        \centering
        \includegraphics[width=0.7\columnwidth]{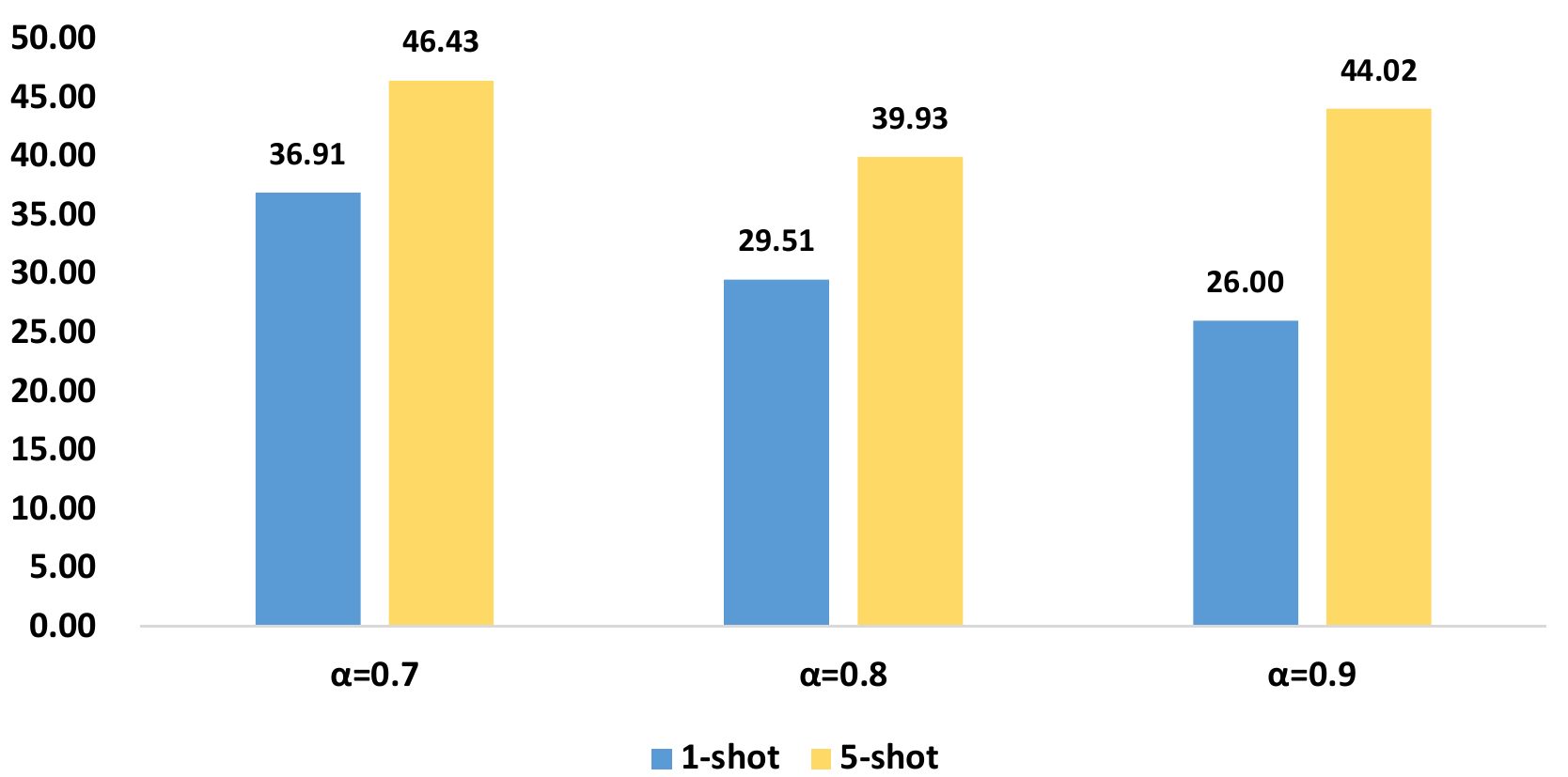}
	\caption{ Micro F1 results with different distance parameter $\alpha$ settings. We fixed the $k=3$,  and $\beta=0.5$. }
	\label{con:parameter_setting_1}
\end{figure}
We also analyze the impact of $\alpha\in\{0.7, 0.8,0.9\}$, which is the threshold to select the positive prototype based on the type description. As Figure~\ref{con:parameter_setting_1} shows, \textsc{W-Procer} achieves the best performance when $\alpha$ is $0.7$. 
When $\alpha>0.8$, too few  cluster centers are considered positive, especially in the 1-shot  situation. 

\subsubsection{Choice of Entity Description}
\label{con:Choise}
\begin{table}[t]
	\centering
  \resizebox{0.5\textwidth}{!}{%
	\renewcommand\arraystretch{1.3}
	\begin{tabular} {l|cc}
	    \toprule 
		\multirow{2}*{}& \multicolumn{2}{c}{NCBI}  \\
		 & 1-shot &5-shot   \\ 
		 \midrule	
		Surface name& 38.85$\pm$3.50 &  39.09$\pm$1.81  \\
		Prototypical instances& 38.50$\pm$4.42 &  40.51$\pm$3.09  \\ 
		Description&  \textbf{38.86$\pm$2.80}&\textbf{40.90$\pm$1.63}   \\ 	
	\bottomrule    
	\end{tabular}
 }
		\caption{Micro F1 scores on NCBI test data with different entity type descriptions.}\label{con:entity_description}
\end{table}

In Table~\ref{con:entity_description},  we compare the \textsc{W-Procer} performance with different entity type descriptions  on the NCBI dataset.  Our final model utilizes the annotated type description,  which outperforms other options: (1) the surface name for each entity type be regarded as the input to the entity type encoder, such as,  in the dataset NCBI, surface name \textit{SpecificDisease} for entity type \textit{SpecificDisease}. (2) In the training progress, we sample the entities for each type as the prototypical instances and use them as input to the entity type encoder. In the 1/5-shot setting, we use 1/5 entities of each type as the input of the type encoder.

\subsubsection{Model Performance with Different Mask Strategies}
\label{con:mask}

\begin{table}[t]
	\centering
 \resizebox{0.5\textwidth}{!}{%
	\renewcommand\arraystretch{1.3}
	\begin{tabular} {l|ccc}
	\toprule 
		\multirow{2}*{}& \multicolumn{3}{c}{NCBI}  \\
		& mask-0 &mask-2   &mask-3 \\ 
		\midrule
		BioBERT& 35.88$\pm$2.55  & 30.16$\pm$2.47  &  26.40$\pm$2.91\\ 
		GatorTron& 37.64$\pm$1.76    & 35.63$\pm$2.96&29.72$\pm$3.40\\ 	
  	Ours&    \textbf{40.90$\pm$1.63} & \textbf{37.29$\pm$1.93}  &\textbf{33.16$\pm$3.24} \\
		\bottomrule 
	\end{tabular}
 }
	\caption{F1 scores on NCBI  test data with different entity mask ratios in the situation of 5-shot.}\label{con: Mask_methods}
\end{table}

In this section, our objective is to examine the influence of various masking rates on the model and determine the continued efficacy of our contrastive learning method on post-masking.
This experiment comprehensively simulated a scenario where numerous unlabeled entities exist within the sentence.

Specifically,  in the 5-shot setting, we introduce two mask Strategies in the training dataset:
\begin{itemize}
    \item \textbf{Mask-2} strategy, where only  two entities per type are present, while the remaining entities are labeled as "O".
    \item \textbf{Mask-3} strategy, where only  three entities per type are present, while the remaining entities are labeled as "O".
\end{itemize}

By the experiments result in Table~\ref{con: Mask_methods},  we have the following observations:  (1) As the mask ratio increases, the model performance will decrease. 
We guess  that this decline is attributed to the excessive amount of unlabeled data, which can adversely affect the model's performance;
(2) we observe that our models  still can get good performance even with significantly high masking rates, which validates our assumption that our contrastive loss with cluttering the \textsc{Outside (O)} labeled tokens aids in improving the discrimination between entities belonging to different labels, even in the presence of unlabeled entities.

\subsubsection{Visualization of Learning Representation}
\label{con:visual}

\begin{figure}[htbp]
    \centering
    \subfigure[]{
        \includegraphics[width=1.11in]{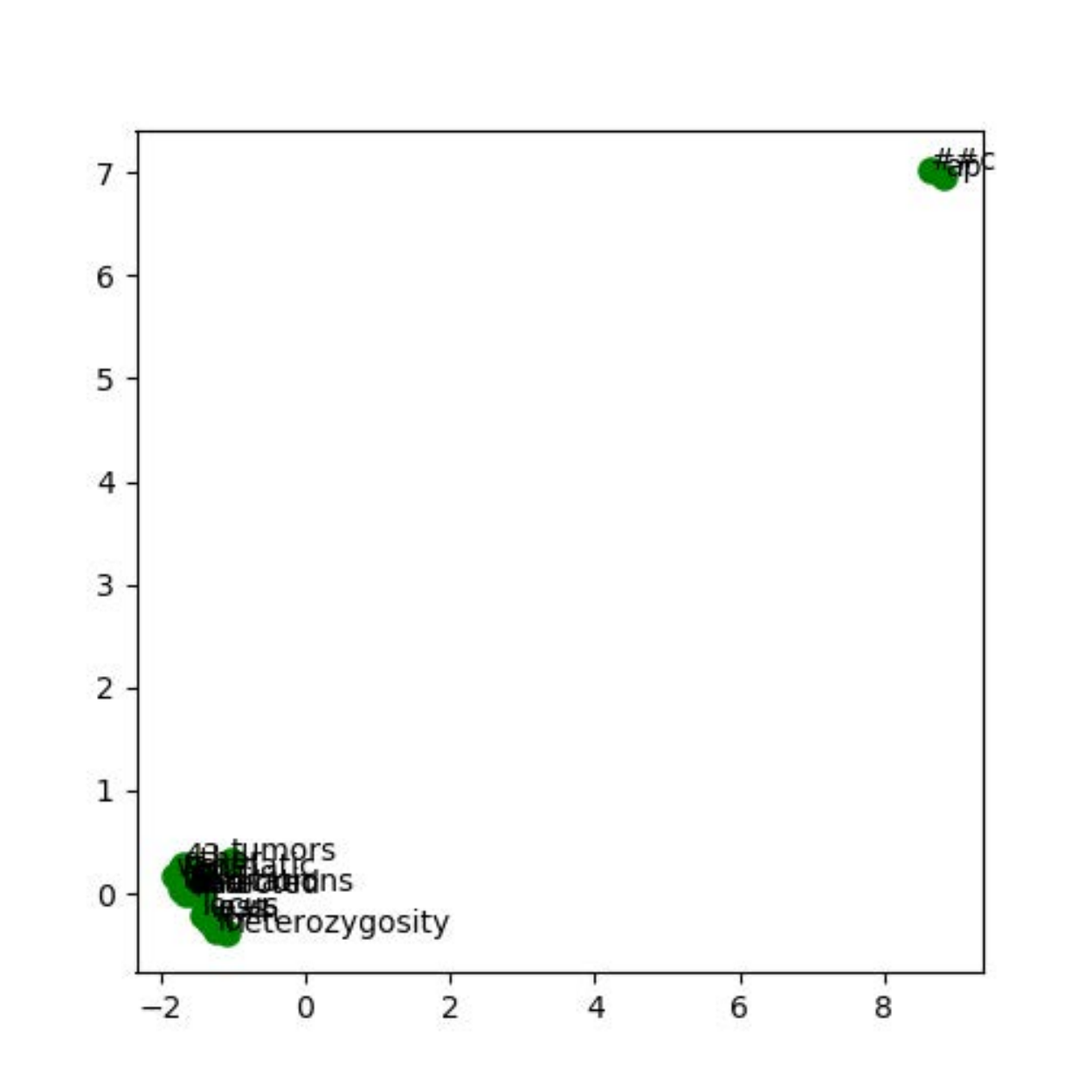}
    }
    \subfigure[]{
	\includegraphics[width=1.11in]{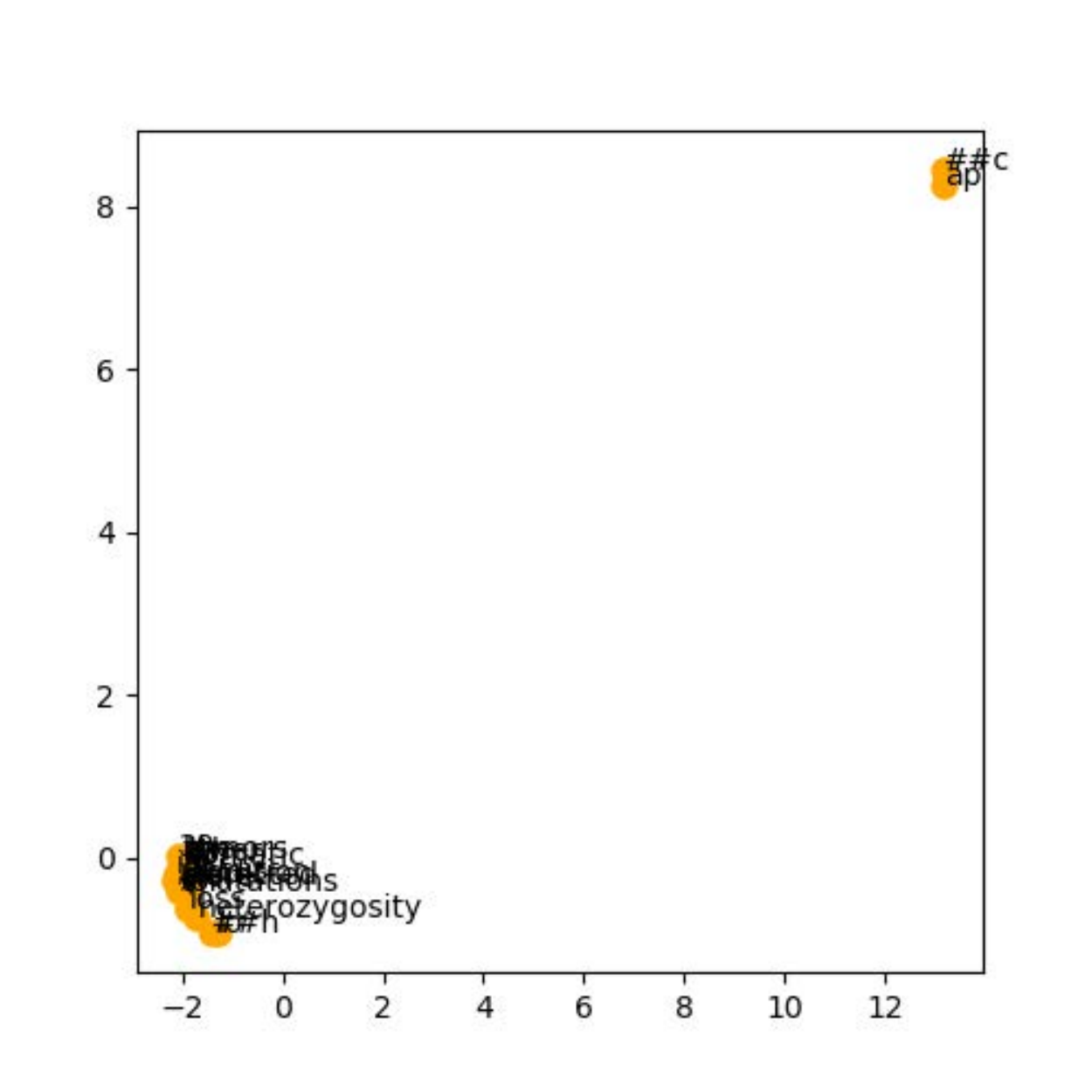}
    }
    \quad    
    \subfigure[]{
    	\includegraphics[width=1.2in]{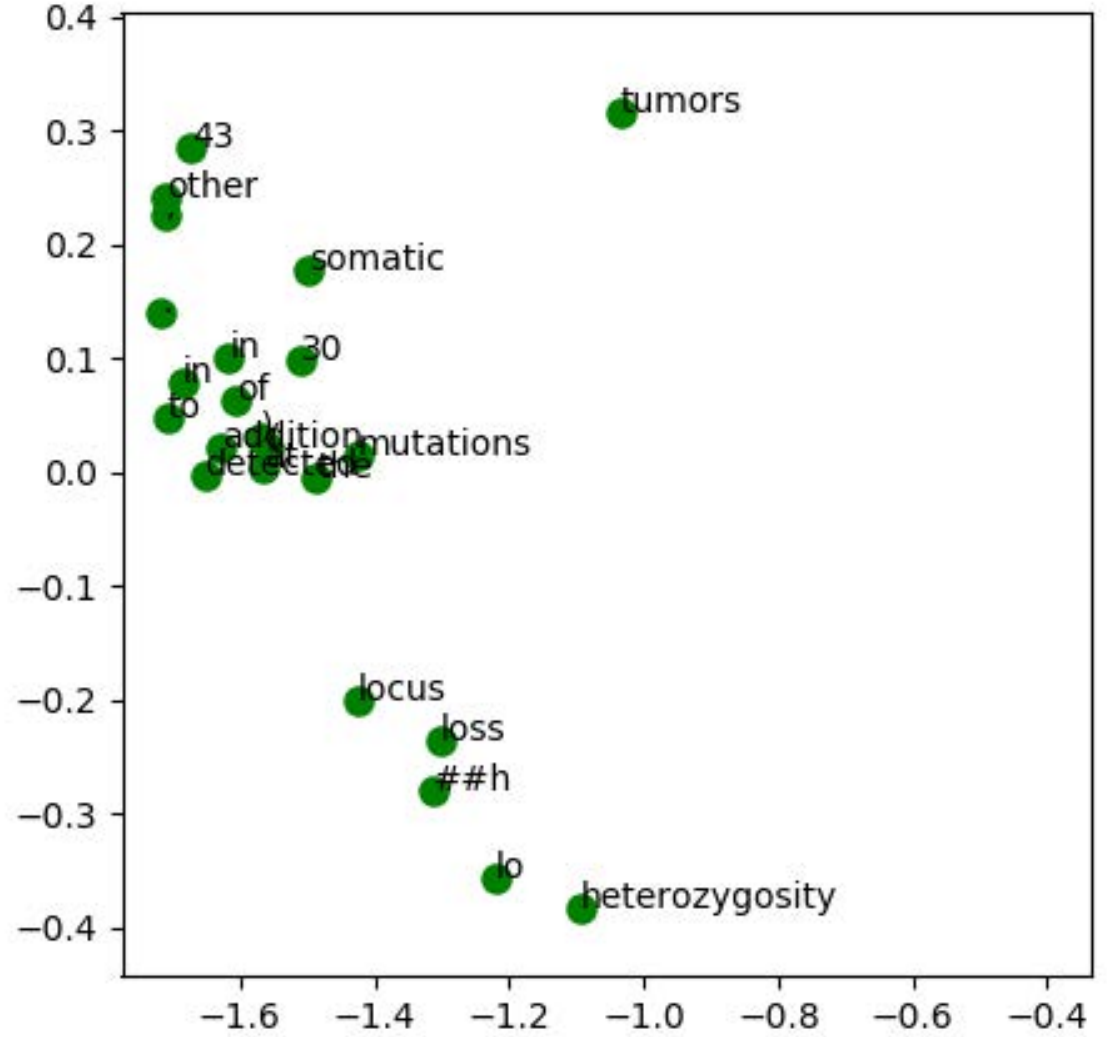}
    }
    \subfigure[]{
	\includegraphics[width=1.2in]{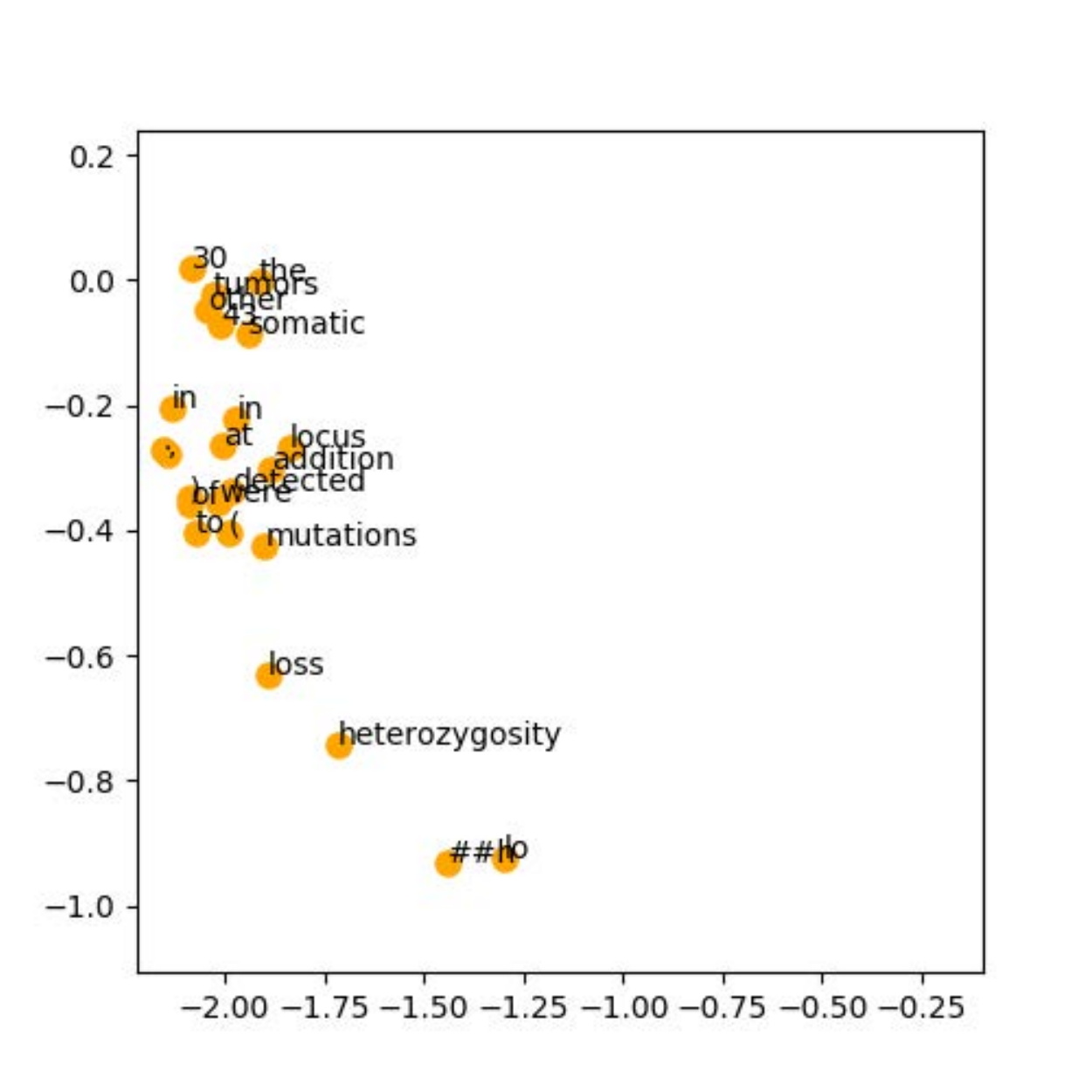}
    }
    \caption{(a)  and (b) refer to \textsc{W-Procer}  and  GatorTron Representation for sentence $S$ separately. (c)  and (d) refer to \textsc{W-Procer}  and  GatorTron Representation for the Outside (O) labeled tokens in the sentence $S$ separately.}
    \label{Visualization_Chart}
\end{figure}

In Figure~\ref{Visualization_Chart}, we visualize the token embedding of an example sentence in the dataset NCBI, $S$="\textit{In addition to loss of heterozygosity (LOH) at the APC locus in 30 tumors, 43 other somatic mutations were detected.
}", its gold label sequence is "\textit{O O O O O O O O O Modifier O O O DiseaseClass O O O O O O O}", the label of the entity \textit{APC} is \textit{Modifier}, the label of the entity \textit{tumors} is \textit{DiseaseClass}.
To assess the model's proficiency in distinguishing labels, we employ the technique \textbf{Mask-3} described in section~\ref{con:mask}  to train the \textsc{W-Procer} and GatorTron. Additionally, we utilize \textsc{W-Procer} and GatorTron to visualize the token embedding on $S$.  Using this masking strategy, in sentence $S$, the label for the word "\textit{tumors}" is replaced with the label "\textit{O}".
Based on the visualizations in Figure~\ref{Visualization_Chart} (a) and (b), it is evident that both models are capable of distinguishing the entity "APC". In Figure~\ref{Visualization_Chart} (c) and (d), the representation learned by the proposed \textsc{W-Procer} exhibits more distinct clusters compared to the representation learned by GatorTron. Notably, the labeled word "tumors" maintains a significant distance from the Outside (O) labeled tokens, even though it is labeled as \textit{O} during the training process. This finding further supports the effectiveness of our prototype and weighted network-based Contrastive Learning (CL) method.

\section{Conclusions and Future Work}
This paper introduces  \textsc{W-Procer}, a weighted prototypical contrastive learning framework for the few-shot medical NER task in the medical domain. It consists of two primary steps. First, positive and negative prototypes are constructed using tokens labeled as \textsc{Outside (O)}. Then, a prototype-based contractive loss is employed along with a weighting network to enhance the model's capability to distinguish the incorrect negative samples from \textsc{Outside (O)} labeled tokens. 
This approach helps attract positive prototypes and repel negative prototypes, ultimately improving the model's distinguish ability. Our model has been proven effective in addressing the Few-shot NER task through experimental results and extensive analysis of three public benchmark datasets. More importantly, our weighted prototypical contrastive learning
framework not only addresses the class collision problem in few-shot medical Named Entity Recognition (NER), but it is also applicable to other medical NLP tasks facing similar challenges. Future research will explore its potential in these domains.
In our study, we utilized the small language model GatorTron as the foundation for our model. While our current framework has shown satisfactory results in addressing the challenges of medical secret data, our future focus will be on developing an innovative model that harnesses the power of open large language models, such as LLaMA, to offer an enhanced solution for handling sensitive data. 

\section*{Acknowledgements}
This work was supported by the National Institutes of Health’s National Center for Complementary and Integrative Health grant number R01AT009457 and National Institute on Aging grant number R01AG078154. The content is solely the responsibility of the authors and does not represent the official views of the National Institutes of Health.

\section{Competing Interests}
The authors declare that there is no conflict of interest.



\section{CRediT authorship contribution statement}

\textbf{Mingchen Li}:  Methodology, Software, Writing – original draft. \textbf{Yang Ye}:   Writing - Review and Editing.  \textbf{Huixue Zhou}:  Software, Writing - Review and Editing. \textbf{Jeremy Yeung}: Software, Writing - Review and Editing. \textbf{Huaiyuan Chu}: Software, Writing - Review and Editing. \textbf{Rui Zhang}:Conceptualization, Methodology, Supervision, Writing – original draft, Writing – review \& editing.

\appendix


 \bibliographystyle{elsarticle-num} 
 \bibliography{anthology}





\end{document}